\documentclass[letterpaper]{article} % DO NOT CHANGE THIS
\usepackage{aaai25}  % DO NOT CHANGE THIS
\usepackage{times}  % DO NOT CHANGE THIS
\usepackage{helvet}  % DO NOT CHANGE THIS
\usepackage{courier}  % DO NOT CHANGE THIS
\usepackage[hyphens]{url}  % DO NOT CHANGE THIS
\usepackage{graphicx} % DO NOT CHANGE THIS
\usepackage{natbib}  % DO NOT CHANGE THIS AND DO NOT ADD ANY OPTIONS TO IT
\usepackage{caption} % DO NOT CHANGE THIS AND DO NOT ADD ANY OPTIONS TO IT
\usepackage{algorithm}
\usepackage{newfloat}
\usepackage{listings}
\DeclareCaptionStyle{ruled}{labelfont=normalfont,labelsep=colon,strut=off} % DO NOT CHANGE THIS
\floatstyle{ruled}
\newfloat{listing}{tb}{lst}{}
\floatname{listing}{Listing}
\usepackage{titletoc} % Needed for creating partial ToC
\usepackage{mathtools}
\usepackage{algorithm}
\usepackage{colortbl}
\usepackage{caption}
\usepackage{subfigure}
\usepackage{algpseudocode}
\usepackage{amssymb}
\usepackage[algo2e]{algorithm2e} % For algorithms % when uncommented creates issue!
\usepackage[textsize=tiny]{todonotes}
\definecolor{revised_color}{RGB}{250 150 0}
\definecolor{removed_material}{RGB}{250 0 50}
\definecolor{unresolved}{RGB}{250 150 0}

\definecolor{navid_color}{RGB}{50 200 100}
\definecolor{revise_color}{RGB}{200 50 150}

\definecolor{lightgray}{gray}{0.9}

\definecolor{comment_color}{RGB}{0 151 57}
\title{Uncertainty separation via ensemble quantile regression}
\author{
    Written by AAAI Press Staff\textsuperscript{\rm 1}\thanks{With help from the AAAI Publications Committee.}\\
    AAAI Style Contributions by Pater Patel Schneider,
    Sunil Issar,\\
    J. Scott Penberthy,
    George Ferguson,
    Hans Guesgen,
    Francisco Cruz\equalcontrib,
    Marc Pujol-Gonzalez\equalcontrib
}
\title{Uncertainty separation via ensemble quantile regression}
\author {
    Navid Ansari,
   Hans-Peter Seidel,
    Vahid Babaei
}
\affiliations {
    Max Planck Institute for Informatics\\
    nansari@mpi-inf.mpg.de, 
    hpseidel@mpi-sb.mpg.de,
    vbabaei@mpi-inf.mpg.de
}

\begin{document}

\maketitle

\begin{abstract}
This paper introduces a novel and scalable framework for uncertainty estimation and separation with applications in data driven modeling in science and engineering tasks where reliable uncertainty quantification is critical. Leveraging an ensemble of quantile regression (E-QR) models, our approach enhances aleatoric uncertainty estimation while preserving the quality of epistemic uncertainty, surpassing competing methods, such as Deep Ensembles (DE) and Monte Carlo (MC) dropout. To address challenges in separating uncertainty types, we propose an algorithm that iteratively improves separation through progressive sampling in regions of high uncertainty. Our framework is scalable to large datasets and demonstrates superior performance on synthetic benchmarks, offering a robust tool for uncertainty quantification in data-driven applications.
\end{abstract}

% \include{sections/Intro}
\section{Introduction}
\label{sec:intro}

In recent years, uncertainty estimation has become an essential aspect of machine learning, especially for applications that demand high reliability in decision-making, such as autonomous driving \cite{kendall2017uncertainties}, and medical diagnosis \cite{lambrou2010reliable, yang2009using}. Accurate uncertainty estimation not only supports better model interpretability but also helps identify areas where models are likely to make errors, ensuring safety in high-stakes environments.

Uncertainty in machine learning is typically categorized into two types: \textit{aleatoric} uncertainty, which originates from inherent noise or variability in the data, and \textit{epistemic} uncertainty, which stems from limitations in the model’s knowledge and can potentially be reduced with additional data \cite{hullermeier2021aleatoric}. Separating these two types of uncertainty is essential in applications like Bayesian optimization \cite{frazier2018tutorial,hernandez2017parallel, shahriari2015taking,ansari2023large}, inverse design \cite{wijaya2024trustmol, ansari2022autoinverse}, and active learning \cite{ren2021survey, settles2009active, kirsch2019batchbald, gal2017deep}, where understanding the nature of uncertainty influences strategic decisions. 

For instance, in Bayesian optimization (BO) \cite{frazier2018tutorial,hernandez2017parallel, shahriari2015taking,ansari2023large} and active learning  \cite{ren2021survey, settles2009active, kirsch2019batchbald, gal2017deep}, focusing on epistemic uncertainty directs computational resources toward regions where additional data could improve model performance.
In engineering, inverse design is a critical task aimed at identifying design parameters that achieve a desired performance. Accurate characterization of uncertainty—both its magnitude and type— is essential for this process, as it enables the identification of designs that are not only optimized for performance but also robust and reliable under real-world conditions \cite{wijaya2024trustmol, ansari2022autoinverse}.

While various methods exist for modeling uncertainty, most classical BO approaches relying on Guassian processes as surrogate models cannot effectively scale with increasing dataset size which is a common theme in engineering problems \cite{wang2016bayesian, hernandez2015probabilistic, snoek2012practical}. For this reason there has been a lot of efforts in replacing GPs with Bayesian neural networks \cite{neal2011mcmc, chen2014stochastic, lakshminarayanan2016simple, gal2016dropout, lee2017deep, wilson2016deep}. However \citet{ansari2023large} showed that most of these BNNs also do not scale very well with the size of the dataset. Among the techniques capable of handling large-scale data are Deep Ensembles (DE) \cite{Lakshminarayanan2017} and Monte Carlo (MC) dropout \cite{Gal2016}. While the former can provide some degree of uncertainty separation MC dropout only predicts the epistemic uncertainty. However, even DE faces limitations in both aspects of uncertainty estimation quality:

\begin{itemize}
    \item \textbf{Localization of uncertainty:} The first task in uncertainty estimation is identifying regions within the input space that may lead to unreliable predictions, regardless of the type of uncertainty. Effective localization aids in detecting areas where model predictions might not remain faithful to reality.

    \item \textbf{Separation of uncertainties:} The next task involves distinguishing between aleatoric and epistemic uncertainty across the input domain. This distinction is particularly valuable in applications such as Bayesian optimization and active learning, where it is beneficial to avoid exploring regions dominated by irreducible aleatoric noise.

\end{itemize}

% In this work, we address both challenges by utilizing an ensemble of quantile regression (E-QR) as an alternative to Deep Ensembles (DE) \ref{hoel2023ensemble, mallick2022deep, tagasovska2018frequentist}. We demonstrate that E-QR provides superior aleatoric uncertainty estimation while being more computation efficient and stable (\navid{ref, table, numbers}) (Section\navid{xxx}). In Section \navid{xxx},
We demonstrate that under certain conditions, existing methods can misidentify uncertainty types and provide inaccurate predictions. To address these issues, we introduce Algorithm \ref{alg:u_separation} designed to enhance accuracy and reliability in uncertainty separation. 

In Section \ref{sec:eval}, we apply DE and E-QR to both a toy problem and a synthetic mechanical problem to illustrate the challenges associated with uncertainty separation. We further demonstrate how Algorithm \ref{alg:u_separation} effectively addresses these challenges, highlighting its advantages in accurately separating and quantifying uncertainty.
%%%%%%%%%%%%%%%%%%%%%%%%%%%%%%%%%%
\section{Related work and background}
\label{sec:related}

Uncertainty estimation in machine learning, particularly in deep learning models, has gained significant attention due to its importance in reliable decision-making. Separating epistemic and aleatoric uncertainty is critical in many applications. Deep Ensembles (DE), introduced by \citet{Lakshminarayanan2017}, provides uncertainty estimates by averaging predictions from independently trained neural networks. While DE can estimate both epistemic and aleatoric uncertainty, it sometimes fails to separate them effectively.

Monte Carlo (MC) dropout approximates Bayesian neural networks by applying dropout during training and inference \cite{Gal2016}. 

Quantile regression has emerged as a promising approach for uncertainty estimation, predicting the upper and lower quantiles of the target distribution instead of point estimates \cite{Koenker1978}. To ensemble such models, \citet{fakoor2023flexible} proposed a paradigm focused on aleatoric uncertainty capture but did not address epistemic uncertainty separation. \citet{tagasovska2018frequentist} leveraged quantile regression ensembles to model epistemic uncertainty in high-dimensional spaces, while \citet{hoel2023ensemble} applied it to reinforcement learning in autonomous driving, demonstrating robustness in safety-critical scenarios. Additionally, \citet{mallick2022deep} showcased its scalability and accuracy in separating uncertainties in spatiotemporal problems.

DE and Ensemble Quantile Regression (E-QR) are superior to MC dropout for epistemic uncertainty estimation, as both train independent models with varying parameters, unlike MC dropout, which uses stochastic variations of a single model \cite{gal2016dropout, lakshminarayanan2016simple, koenker2001quantile}. While DE uses sub-networks trained with Negative Log Likelihood (NLL) loss to model aleatoric uncertainty, E-QR leverages pinball loss to predict quantiles \cite{romano2019conformalized}.

E-QR is simpler and more stable to train than DE. Pinball loss is less sensitive to initialization and learning rate compared to NLL loss, which is prone to gradient instabilities and overfitting on small datasets \cite{moustakides2019training, streit1994maximum}. DE requires a two-step process—optimizing primary predictions (\(\mu\)) followed by training uncertainty (\(\sigma\)) heads with NLL loss—doubling its computational effort compared to E-QR, which learns quantile predictions in a single step \cite{zhang2020novel, ansari2022autoinverse}. Moreover, E-QR provides superior aleatoric uncertainty predictions by directly modeling quantile intervals, avoiding parametric assumptions required in DE \cite{barron2019general}.

\section{Methodology}
In Section \ref{sec:separation}, we discuss the shortcomings of scalable Bayesian Neural Networks (BNNs) in uncertainty separation and demonstrate how Algorithm \ref{alg:u_separation} can be used to achieve robust uncertainty separation.
% In Section \ref{sec:E_QR}, we describe the ensembling of quantile regression models as an alternative to DE and MC dropout.

\subsection{Challenges in uncertainty separation}
\label{sec:challenges}
\textbf{Leakage of aleatoric uncertainty into epistemic uncertainty}

Although epistemic uncertainty is intended to capture only uncertainty due to limited knowledge, in practice, it may inadvertently include aleatoric uncertainty when data is insufficient. This occurs because each subnetwork in an ensemble model is trained on a subsample of the data. If these subsamples are too small, subnetworks may overfit to their specific data rather than generalizing between data points, as expected with an L2 loss function. As a result, aleatoric uncertainty can ``leak" into the epistemic uncertainty estimates. This phenomenon is common in both Deep Ensembles and Ensemble Quantile Regression methods.

\begin{figure}
\centering
\subfigure[]{\label{fig:aleatoric_to_epistemic_leak}\includegraphics[width=0.20\textwidth]{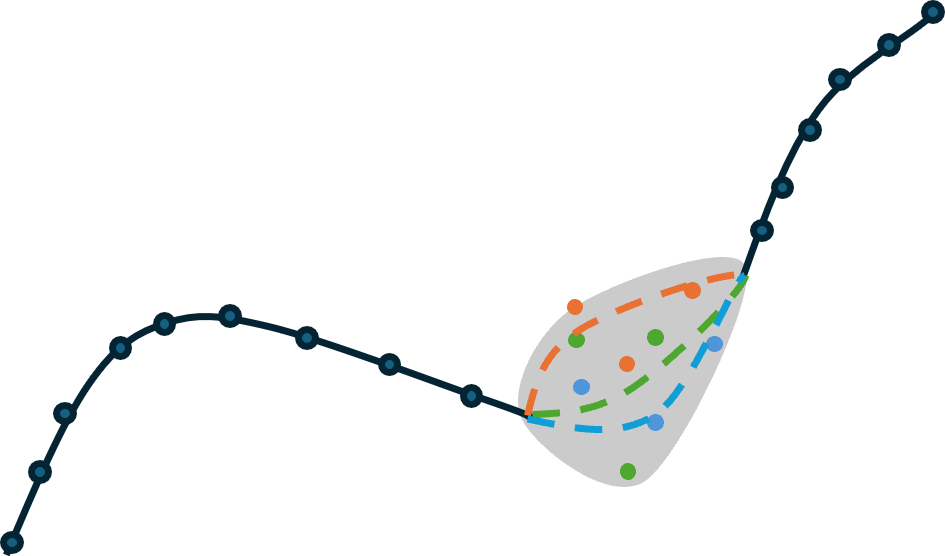}}
\subfigure[]{\label{fig:epistemic_to_aleatoric_leak}\includegraphics[width=0.20\textwidth]{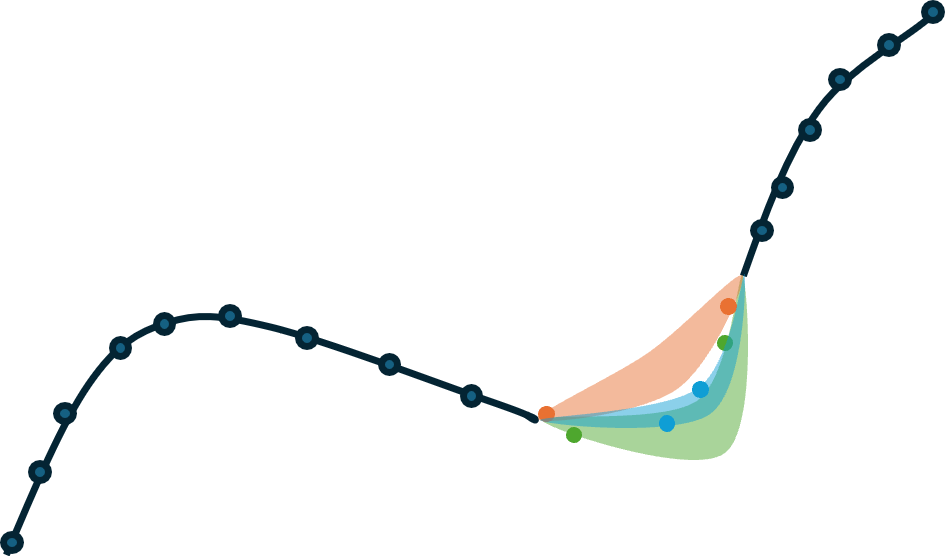}}
\caption{On the left figure, we observe how the lack of data in a region with aleatoric uncertainty causes each sub-network to fit the noise differently, as they only access small subsets of the data. This overfitting results in the false reporting of epistemic uncertainty where none exists. On the right figure, we see that in regions lacking sufficient data, the fits for aleatoric uncertainty become unreliable, leading to incorrect report of aleatoric uncertainty.
}
\label{fig:leak}
\end{figure}
\textbf{Leakage of epistemic uncertainty in aleatoric uncertainty.}

All models studied here rely on fitting mechanisms to model aleatoric uncertainty. DE uses the Negative Log Likelihood (NLL) loss, while E-QR employs the pinball loss to fit upper and lower quantiles. As a result, the accuracy of these predictions depends heavily on the availability of high-quality and sufficient data.
In regions with high epistemic uncertainty, the fit of aleatoric uncertainty estimates can become arbitrarily. This issue is exacerbated by the use of bagging in sub-networks, where only a subsample of data is visible to models attempting to learn aleatoric uncertainty.

Knowing that the root cause of both problems is the lack of data, we propose a progressive sampling strategy in the next section. This strategy focuses on regions where uncertainty is detected, but the type of uncertainty (aleatoric or epistemic) remains unclear. By progressively acquiring additional data in these regions, the model can refine its predictions and better separate the two types of uncertainty.

\subsection{Reliable separation of epistemic and aleatoric uncertainty}
\label{sec:separation}
To reliably separate aleatoric and epistemic uncertainties, we first identify regions where uncertainty is suspected and create a comprehensive uncertainty map by normalizing and combining all available uncertainty maps. In cases with multiple outputs, we focus on the common regions across each output's uncertainty map, as these shared uncertainties are more likely to originate from the data rather than local misfitings specific to individual outputs.

Once this common uncertainty map is established, we gather additional data in the uncertain regions and retrain the models. Iteratively repeating this process diminishes regions of epistemic uncertainty while areas of aleatoric uncertainty remain unchanged. By performing a logical XOR operation between the final uncertainty map and the initial one, we can isolate the initial epistemic uncertainty present in our initial dataset.
Algorithm \ref{alg:u_separation} presents the complete procedure.

%%%%%%%%%%%%%%%%%%%%%%%%%% Algorithm %%%%%%%%%%%%%%%%%%%%%%%%%%%
\begin{algorithm}
\scriptsize % Adjust the font size
\textbf{Input}

$(X, Y)^0$  \hspace{0.4cm}  \textcolor{comment_color}{\tcp{Initial data set}}\
$Q$  \hspace{0.4cm} \textcolor{comment_color}{\tcp{Number of iterations of the main algorithm}}\
$O$  \hspace{0.4cm} \textcolor{comment_color}{\tcp{Number of outputs}}\
$\Phi$  \hspace{0.4cm} \textcolor{comment_color}{\tcp{ Native Forward Process, e.g., a simulation}}\
$T$  \hspace{0.4cm} \textcolor{comment_color}{\tcp{Threshold value for Binarizing the uncertainty map}}\
\textbf{Output}

$U_{E}$, $D_{UE}$ , $P_{UE}$ \hspace{0.4cm} \textcolor{comment_color}{\tcp{Position and value of the separated \textbf{epistemic} regions.}}\
$U_{A}$, $D_{UA}$ , $P_{UA}$ \hspace{0.4cm} \textcolor{comment_color}{\tcp{Position and value of the separated \textbf{aleatoric} regions.}}\

\Begin{
    $data set \gets (\mathbf{X^{0}}$,\: $\mathbf{Y^{0}})$\
    
    $f_{BNN}^{0} \xLeftarrow[\text{}]{\text{train}} dataset$ \textcolor{comment_color}{\tcp{Train the BNN surrogate.}}

    $U_A, U_E \gets f_{BNN}^{0}$ \textcolor{comment_color}{\tcp{calculate the uncertainties from the surrogate.}}

     $\min{U_E}, \max{U_E}, \min{U_A}, \max{U_A} \gets \text{\textit{MIN-MAX}}(U_E, U_A)$ 
     \textcolor{comment_color}{\tcp{Extracting the min-max of the uncertainties to be used for scaling.}}
    \For{$i\gets1$ \KwTo $Q$}{
        \For{$ j \gets1$ \KwTo $O$}{

            $\overline{U_E}, \overline{U_A} \gets \textit{Normalizer}(\min{U_E}, \max{U_E}, \min{U_A}, \max{U_A})$ 
              \textcolor{comment_color}{\tcp{Scaling and normalizing both total uncertainty and epistemic uncertainty calculated from the original dataset.}}
            $ \overline{U_{total}^j} = \overline{U_E} + \overline{U_A}$ 
              \textcolor{comment_color}{\tcp{ Adding both normalized uncertainties to make sure all the uncertain regions are captured.}}
            $ \overline{U_{total}^{total}} = \overline{U_{total}^{total}} \times \overline{U_{total}^j}$ 
              \textcolor{comment_color}{\tcp{Multiplying all the total uncertainty maps for all outputs to make sure we only keep the ones that are mutual.}}
            $ (X, Y)^i \gets \textit{Binarize}(\overline{U_{total}^{total}}, T)$
              \textcolor{comment_color}{\tcp{Binarizing the uncertainty map to create a mask for extracting the next batch of data $(X, Y)^i$.}}
              $data set \gets (X,Y)^i$
              \textcolor{comment_color}{\tcp{Append new data to the old.}}\
              $f_{BNN}^{i} \xLeftarrow[\text{}]{\text{train}} data set$  \textcolor{comment_color}{\tcp{Train the BNN surrogate.}}\
              $U_A^i, U_E^i \gets f_{BNN}^{i}$ \textcolor{comment_color}{\tcp{calculate the uncertainties from the surrogate.}}       
         } 
        }
        $U_{A}, D_{UA} , P_{UA} \gets \overline{U_{total}^{total}}$ \textcolor{comment_color}{\tcp{After sufficient number of iterations the total uncertainty map only contains aleatoric uncertainty.}}
        $U_{A}, D_{UA} , P_{UA} \gets  \overline{U_{total}^{total}} \oplus \overline{U_{total}^{0}}$ \textcolor{comment_color}{\tcp{By comparing the final total uncertainty with the initial one we can determine the epistemic uncertain regions of the initial data.}} 
        
\caption{Uncertainty separation.}
\label{alg:u_separation}
 }
\end{algorithm}

\section{Evaluation}
\label{sec:eval}
\subsection{Experiment setup}
\label{sec:experiment_setup}

\paragraph{\textbf{Toy}}
In this experiment, we aim to fit the function 
\[
y_1 = \sin \left(\sqrt{x_1^2 + x_2^2}\right), y_2 = x \cdot \cos\left(\sqrt{x_1^2 + x_2^2}\right) \cdot \cos(x_2)
\]

We introduce 4 specific regions in the input space two on the top lacking data, and two on the bottom polluted with irreducible random noise modeled as $\mathcal{N}(0, 0.3)$ (Figure \ref{fig:alg_toy}).

\paragraph{\textbf{Multi-joint robot}}
In this problem, a robotic arm with four rotatable joints and an adjustable base position on the wall (\(x \in \mathbb{R}^5\)) is considered. The goal is to predict the final 2D position of the arm's tip (\(y \in \mathbb{R}^2\)) given its joint angles \cite{ardizzone2018analyzing}. To evaluate uncertainty separation, we conduct an experiment where the behavior of one joint is excluded from the dataset to test whether our model can recover the missing information and determine the type of uncertainty. 
% Additionally, in a complementary experiment described in \navid{appendix}, we analyze a more complex scenario where one joint is injected with random noise while another joint's behavior is excluded, demonstrating the algorithm's capability in separating uncertainties under challenging conditions.

% In this problem we have a robotic arm which has 4 rotatable joints and we can also adjust its base position on the wall $x \in \mathcal{R}^5$ (Figure \ref{fig:robotic_arm}).
% % 
% The goal is to find the final 2D position of the tip of the robotic arm given its joint angles $y \in \mathcal{R}^2$ \cite{ardizzone2018analyzing}. 
% %
% We run this experiment by not showing the behavior of another joint to the model to see if we can recover and determine the type of the uncertainty.
% %
% We also run a comjplementary experiment in \navid{appendix } in a more complicated setting where one of the joints is injected with random noise and another joint's bahaviour is excluded from the data set and not shown to the model to see how our algorithm is capabe of separating these uncertainties.
%
\begin{figure}
    \centering        
    \includegraphics[width=0.3\textwidth]{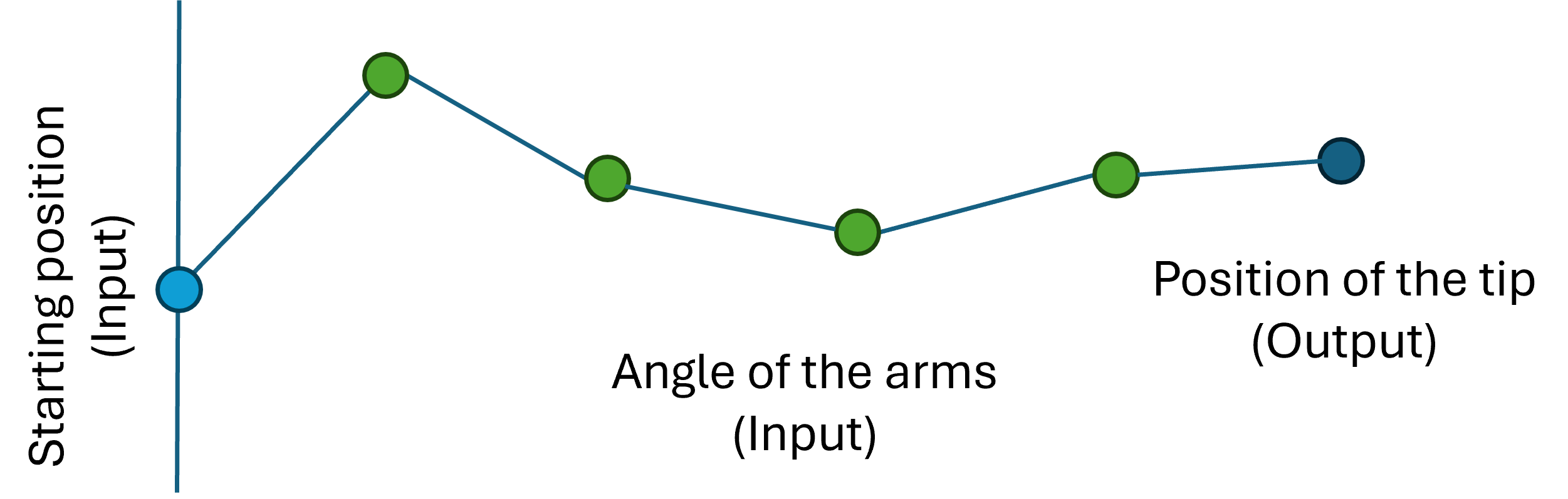}
    \caption{Multi joint robotic arm with a moving base and 4 rotating joints. The goal is to train a model that can predict the 2D position of the tip of the arm.} 
    \label{fig:robotic_arm}
\end{figure}

\subsection{Aleatoric uncertainty leak into epistemic uncertainty prediction}
\begin{figure}[t]
    \centering        
    \includegraphics[width=0.5\textwidth]{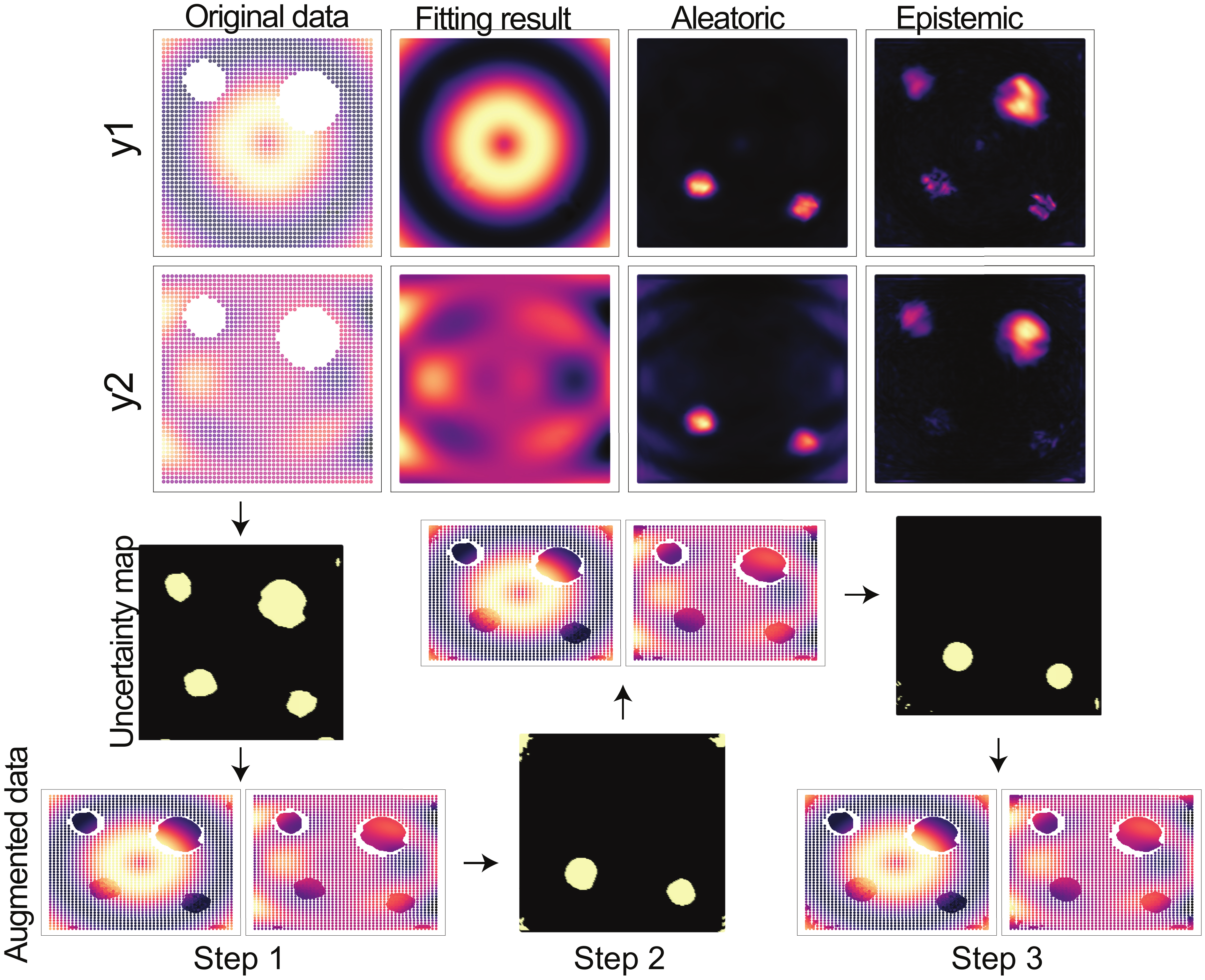}
    \caption{In the top figure, from left to right, we present the original training data, the model’s predictions, and the aleatoric and epistemic uncertainty maps. The first row corresponds to the first output, while the second row corresponds to the second output of the model. The epistemic uncertainty map highlights four regions: two caused by a lack of data and two influenced by the leak of random noise. To achieve accurate separation, we apply Algorithm \ref{alg:u_separation}. After two iterations, only the regions with aleatoric uncertainty remain in the uncertainty map, confirming that the vanished uncertain areas were indeed epistemic. Note that the white dots are due to the low density of the training samples.}
    \label{fig:alg_toy}
\end{figure}
\begin{figure}[t]
    \centering        
    \includegraphics[width=0.5\textwidth]{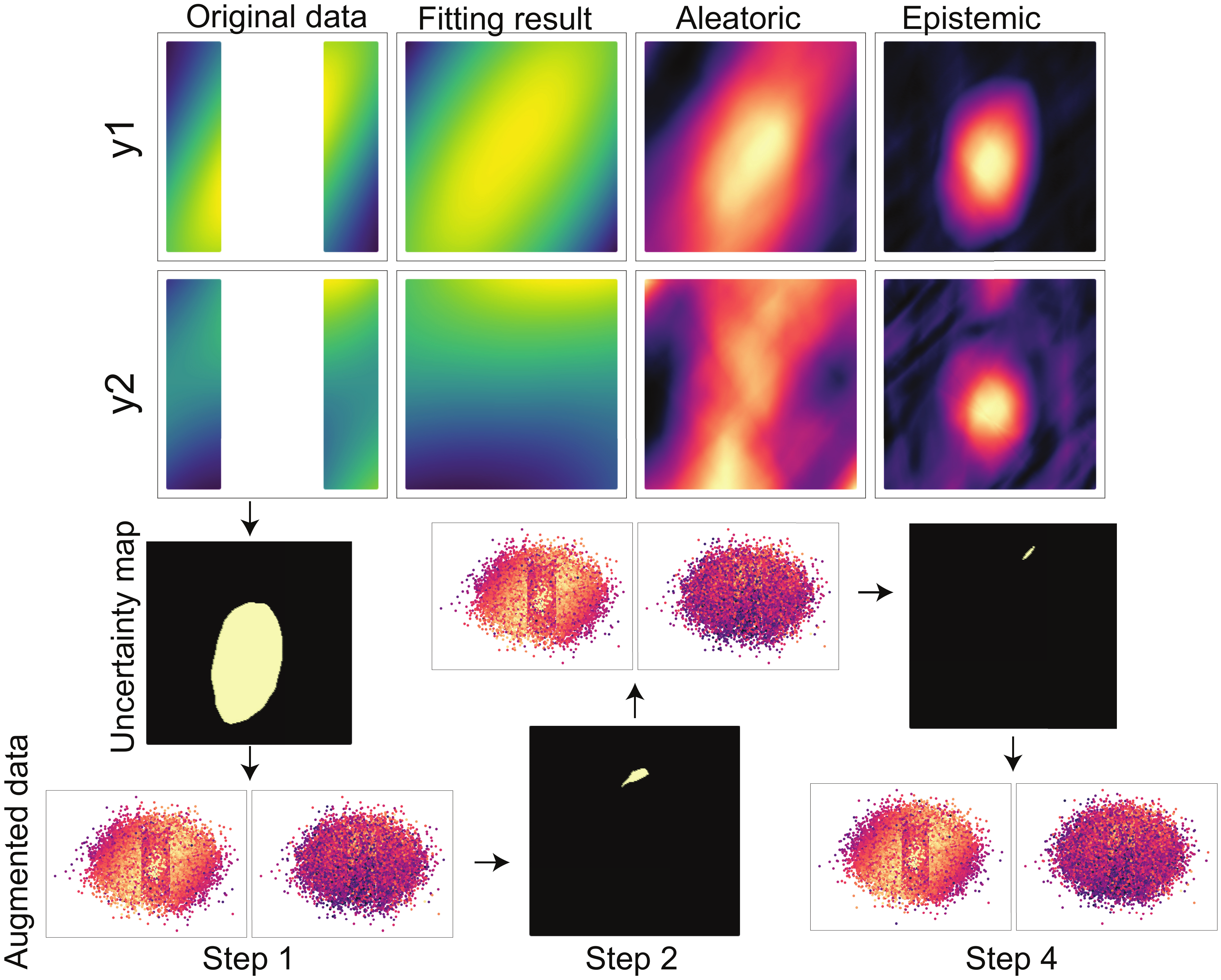}
    \caption{This experiment highlights the leakage of epistemic uncertainty into aleatoric uncertainty. The plot illustrates the cross-section of two out of four rotating joints and their effect on the 2D position of the tip. From left to right, we present the original training data, the model’s predictions, and the aleatoric and epistemic uncertainty maps. Since no random noise is injected into this problem, we expect to observe only epistemic uncertainty. However, the aleatoric uncertainty map incorrectly reflects leaked epistemic uncertainty. Applying Algorithm \ref{alg:u_separation} for four iterations resolves this issue, as the uncertainty vanishes when the uncertain regions are filled with additional data, confirming that the observed uncertainty was indeed epistemic.}
    \label{fig:alg_robot}
\end{figure}

Figure \ref{fig:alg_toy} illustrates the aleatoric and epistemic uncertainty calculated by the E-QR model for both outputs for the toy experiment. Notably, aleatoric uncertainty appears to leak into the epistemic uncertainty plot.

Using Algorithm \ref{alg:u_separation}, we can generate an uncertainty map. By iteratively focusing on uncertain regions and selectively filling them with additional data, the true aleatoric uncertainty regions recovers accurately in one iteration.

\subsection{Epistemic uncertainty leak into aleatoric
uncertainty prediction}

In Figure \ref{fig:alg_robot}, the robotic arm is not augmented with aleatoric noise; however, the behavior of one joint is excluded from the dataset over a range of angles. This setup induces epistemic uncertainty, as the model lacks information about the excluded range. While we expect the model to predict only epistemic uncertainty, the figure shows a leakage of epistemic uncertainty into the aleatoric uncertainty predictions.

Algorithm \ref{alg:u_separation} addresses this issue, identifying the uncertainty as epistemic after four iterations. By adding data to the uncertain regions, the uncertainty is completely resolved, confirming that it was indeed epistemic.

% In Figure \ref{fig:alg_robot} the robot is not augmented with aleatoric noise only for one of the joints we did not show its behavior to the model for a range of angles. We expect our model to predict only epistemic uncertainty but as evident from the figure epistemic uncertainty has leacked into the aleatoric uncertainty prediction.

% Never the less Algorithm \ref{alg:u_separation} after one iteration determins that the uncertainty is indeed epistemic and completely disapears by adding more data in the uncertain regions.

\section{Conclusion}
This work introduces a novel framework for uncertainty separation using Ensemble Quantile Regression (E-QR), addressing the challenges of uncertainty leakage that lead existing methods to erroneous separations. By leveraging E-QR and Algorithm \ref{alg:u_separation}, we achieve robust separation of aleatoric and epistemic uncertainties while mitigating leakage issues. The proposed method is computationally efficient, scalable to large datasets, and validated through experiments on synthetic benchmarks. These results establish our framework as a reliable tool for uncertainty separation in scientific and engineering applications.

\bibliography{aaai25}

\begin{thebibliography}{37}
\providecommand{\natexlab}[1]{#1}

\bibitem[{Ansari et~al.(2023)Ansari, Javanmardi, H{\"u}llermeier, Seidel, and Babaei}]{ansari2023large}
Ansari, N.; Javanmardi, A.; H{\"u}llermeier, E.; Seidel, H.-P.; and Babaei, V. 2023.
\newblock Large-Batch, Iteration-Efficient Neural Bayesian Design Optimization.
\newblock \emph{arXiv preprint arXiv:2306.01095}.

\bibitem[{Ansari et~al.(2022)Ansari, Seidel, Vahidi~Ferdowsi, and Babaei}]{ansari2022autoinverse}
Ansari, N.; Seidel, H.-P.; Vahidi~Ferdowsi, N.; and Babaei, V. 2022.
\newblock Autoinverse: Uncertainty aware inversion of neural networks.
\newblock \emph{Advances in Neural Information Processing Systems}, 35: 8675--8686.

\bibitem[{Ardizzone et~al.(2019)Ardizzone, Kruse, Rother, and Köthe}]{ardizzone2018analyzing}
Ardizzone, L.; Kruse, J.; Rother, C.; and Köthe, U. 2019.
\newblock Analyzing Inverse Problems with Invertible Neural Networks.
\newblock In \emph{International Conference on Learning Representations}.

\bibitem[{Barron(2019)}]{barron2019general}
Barron, J.~T. 2019.
\newblock A general and adaptive robust loss function.
\newblock In \emph{Proceedings of the IEEE/CVF conference on computer vision and pattern recognition}, 4331--4339.

\bibitem[{Chen, Fox, and Guestrin(2014)}]{chen2014stochastic}
Chen, T.; Fox, E.; and Guestrin, C. 2014.
\newblock Stochastic gradient hamiltonian monte carlo.
\newblock In \emph{International conference on machine learning}, 1683--1691. PMLR.

\bibitem[{Fakoor et~al.(2023)Fakoor, Kim, Mueller, Smola, and Tibshirani}]{fakoor2023flexible}
Fakoor, R.; Kim, T.; Mueller, J.; Smola, A.~J.; and Tibshirani, R.~J. 2023.
\newblock Flexible model aggregation for quantile regression.
\newblock \emph{Journal of Machine Learning Research}, 24(162): 1--45.

\bibitem[{Frazier(2018)}]{frazier2018tutorial}
Frazier, P.~I. 2018.
\newblock A tutorial on Bayesian optimization.
\newblock \emph{arXiv preprint arXiv:1807.02811}.

\bibitem[{Gal and Ghahramani(2016{\natexlab{a}})}]{gal2016dropout}
Gal, Y.; and Ghahramani, Z. 2016{\natexlab{a}}.
\newblock Dropout as a bayesian approximation: Representing model uncertainty in deep learning.
\newblock In \emph{international conference on machine learning}, 1050--1059. PMLR.

\bibitem[{Gal and Ghahramani(2016{\natexlab{b}})}]{Gal2016}
Gal, Y.; and Ghahramani, Z. 2016{\natexlab{b}}.
\newblock Dropout as a Bayesian Approximation: Representing Model Uncertainty in Deep Learning.
\newblock In \emph{Proceedings of the 33rd International Conference on Machine Learning (ICML)}, 1050--1059.

\bibitem[{Gal, Islam, and Ghahramani(2017)}]{gal2017deep}
Gal, Y.; Islam, R.; and Ghahramani, Z. 2017.
\newblock Deep bayesian active learning with image data.
\newblock In \emph{International Conference on Machine Learning}, 1183--1192. PMLR.

\bibitem[{Hern{\'a}ndez-Lobato and Adams(2015)}]{hernandez2015probabilistic}
Hern{\'a}ndez-Lobato, J.~M.; and Adams, R. 2015.
\newblock Probabilistic backpropagation for scalable learning of bayesian neural networks.
\newblock In \emph{International conference on machine learning}, 1861--1869. PMLR.

\bibitem[{Hern{\'a}ndez-Lobato et~al.(2017)Hern{\'a}ndez-Lobato, Requeima, Pyzer-Knapp, and Aspuru-Guzik}]{hernandez2017parallel}
Hern{\'a}ndez-Lobato, J.~M.; Requeima, J.; Pyzer-Knapp, E.~O.; and Aspuru-Guzik, A. 2017.
\newblock Parallel and distributed Thompson sampling for large-scale accelerated exploration of chemical space.
\newblock In \emph{International conference on machine learning}, 1470--1479. PMLR.

\bibitem[{Hoel, Wolff, and Laine(2023)}]{hoel2023ensemble}
Hoel, C.-J.; Wolff, K.; and Laine, L. 2023.
\newblock Ensemble quantile networks: Uncertainty-aware reinforcement learning with applications in autonomous driving.
\newblock \emph{IEEE Transactions on Intelligent Transportation Systems}, 24(6): 6030--6041.

\bibitem[{H{\"u}llermeier and Waegeman(2021)}]{hullermeier2021aleatoric}
H{\"u}llermeier, E.; and Waegeman, W. 2021.
\newblock Aleatoric and epistemic uncertainty in machine learning: An introduction to concepts and methods.
\newblock \emph{Machine learning}, 110(3): 457--506.

\bibitem[{Kendall and Gal(2017)}]{kendall2017uncertainties}
Kendall, A.; and Gal, Y. 2017.
\newblock What uncertainties do we need in bayesian deep learning for computer vision?
\newblock \emph{Advances in neural information processing systems}, 30.

\bibitem[{Kirsch, Van~Amersfoort, and Gal(2019)}]{kirsch2019batchbald}
Kirsch, A.; Van~Amersfoort, J.; and Gal, Y. 2019.
\newblock Batchbald: Efficient and diverse batch acquisition for deep bayesian active learning.
\newblock \emph{Advances in neural information processing systems}, 32.

\bibitem[{Koenker and Bassett(1978)}]{Koenker1978}
Koenker, R.; and Bassett, G. 1978.
\newblock Regression Quantiles.
\newblock \emph{Econometrica}, 46(1): 33--50.

\bibitem[{Koenker and Hallock(2001)}]{koenker2001quantile}
Koenker, R.; and Hallock, K.~F. 2001.
\newblock Quantile regression.
\newblock \emph{Journal of economic perspectives}, 15(4): 143--156.

\bibitem[{Lakshminarayanan, Pritzel, and Blundell(2016)}]{lakshminarayanan2016simple}
Lakshminarayanan, B.; Pritzel, A.; and Blundell, C. 2016.
\newblock Simple and scalable predictive uncertainty estimation using deep ensembles.
\newblock \emph{arXiv preprint arXiv:1612.01474}.

\bibitem[{Lakshminarayanan, Pritzel, and Blundell(2017)}]{Lakshminarayanan2017}
Lakshminarayanan, B.; Pritzel, A.; and Blundell, C. 2017.
\newblock Simple and Scalable Predictive Uncertainty Estimation using Deep Ensembles.
\newblock In \emph{Advances in Neural Information Processing Systems (NeurIPS)}, 6402--6413.

\bibitem[{Lambrou, Papadopoulos, and Gammerman(2010)}]{lambrou2010reliable}
Lambrou, A.; Papadopoulos, H.; and Gammerman, A. 2010.
\newblock Reliable confidence measures for medical diagnosis with evolutionary algorithms.
\newblock \emph{IEEE Transactions on Information Technology in Biomedicine}, 15(1): 93--99.

\bibitem[{Lee et~al.(2017)Lee, Bahri, Novak, Schoenholz, Pennington, and Sohl-Dickstein}]{lee2017deep}
Lee, J.; Bahri, Y.; Novak, R.; Schoenholz, S.~S.; Pennington, J.; and Sohl-Dickstein, J. 2017.
\newblock Deep neural networks as gaussian processes.
\newblock \emph{arXiv preprint arXiv:1711.00165}.

\bibitem[{Mallick, Balaprakash, and Macfarlane(2022)}]{mallick2022deep}
Mallick, T.; Balaprakash, P.; and Macfarlane, J. 2022.
\newblock Deep-ensemble-based uncertainty quantification in spatiotemporal graph neural networks for traffic forecasting.
\newblock \emph{arXiv preprint arXiv:2204.01618}.

\bibitem[{Moustakides and Basioti(2019)}]{moustakides2019training}
Moustakides, G.~V.; and Basioti, K. 2019.
\newblock Training neural networks for likelihood/density ratio estimation.
\newblock \emph{arXiv preprint arXiv:1911.00405}.

\bibitem[{Neal et~al.(2011)}]{neal2011mcmc}
Neal, R.~M.; et~al. 2011.
\newblock MCMC using Hamiltonian dynamics.
\newblock \emph{Handbook of markov chain monte carlo}, 2(11): 2.

\bibitem[{Ren et~al.(2021)Ren, Xiao, Chang, Huang, Li, Gupta, Chen, and Wang}]{ren2021survey}
Ren, P.; Xiao, Y.; Chang, X.; Huang, P.-Y.; Li, Z.; Gupta, B.~B.; Chen, X.; and Wang, X. 2021.
\newblock A survey of deep active learning.
\newblock \emph{ACM computing surveys (CSUR)}, 54(9): 1--40.

\bibitem[{Romano, Patterson, and Candes(2019)}]{romano2019conformalized}
Romano, Y.; Patterson, E.; and Candes, E. 2019.
\newblock Conformalized quantile regression.
\newblock \emph{Advances in neural information processing systems}, 32.

\bibitem[{Settles(2009)}]{settles2009active}
Settles, B. 2009.
\newblock Active learning literature survey.

\bibitem[{Shahriari et~al.(2015)Shahriari, Swersky, Wang, Adams, and De~Freitas}]{shahriari2015taking}
Shahriari, B.; Swersky, K.; Wang, Z.; Adams, R.~P.; and De~Freitas, N. 2015.
\newblock Taking the human out of the loop: A review of Bayesian optimization.
\newblock \emph{Proceedings of the IEEE}, 104(1): 148--175.

\bibitem[{Snoek, Larochelle, and Adams(2012)}]{snoek2012practical}
Snoek, J.; Larochelle, H.; and Adams, R.~P. 2012.
\newblock Practical bayesian optimization of machine learning algorithms.
\newblock \emph{Advances in neural information processing systems}, 25.

\bibitem[{Streit and Luginbuhl(1994)}]{streit1994maximum}
Streit, R.~L.; and Luginbuhl, T.~E. 1994.
\newblock Maximum likelihood training of probabilistic neural networks.
\newblock \emph{IEEE Transactions on neural networks}, 5(5): 764--783.

\bibitem[{Tagasovska and Lopez-Paz(2018)}]{tagasovska2018frequentist}
Tagasovska, N.; and Lopez-Paz, D. 2018.
\newblock Frequentist uncertainty estimates for deep learning.
\newblock \emph{arXiv preprint arXiv:1811.00908}.

\bibitem[{Wang et~al.(2016)Wang, Hutter, Zoghi, Matheson, and De~Feitas}]{wang2016bayesian}
Wang, Z.; Hutter, F.; Zoghi, M.; Matheson, D.; and De~Feitas, N. 2016.
\newblock Bayesian optimization in a billion dimensions via random embeddings.
\newblock \emph{Journal of Artificial Intelligence Research}, 55: 361--387.

\bibitem[{Wijaya et~al.(2024)Wijaya, Ansari, Seidel, and Babaei}]{wijaya2024trustmol}
Wijaya, K.~T.; Ansari, N.; Seidel, H.-P.; and Babaei, V. 2024.
\newblock TrustMol: Trustworthy Inverse Molecular Design via Alignment with Molecular Dynamics.
\newblock \emph{arXiv preprint arXiv:2402.16930}.

\bibitem[{Wilson et~al.(2016)Wilson, Hu, Salakhutdinov, and Xing}]{wilson2016deep}
Wilson, A.~G.; Hu, Z.; Salakhutdinov, R.; and Xing, E.~P. 2016.
\newblock Deep kernel learning.
\newblock In \emph{Artificial intelligence and statistics}, 370--378. PMLR.

\bibitem[{Yang et~al.(2009)Yang, Wang, Mi, Lin, and Cai}]{yang2009using}
Yang, F.; Wang, H.-z.; Mi, H.; Lin, C.-d.; and Cai, W.-w. 2009.
\newblock Using random forest for reliable classification and cost-sensitive learning for medical diagnosis.
\newblock \emph{BMC bioinformatics}, 10: 1--14.

\bibitem[{Zhang(2020)}]{zhang2020novel}
Zhang, L. 2020.
\newblock A Novel Penalized Log-likelihood Function for Class Imbalance Problem.

\end{thebibliography}

\end{document}